\documentclass[letterpaper, 10 pt, conference]{ieeeconf}  

\IEEEoverridecommandlockouts                              

\usepackage{graphicx}
\usepackage{xcolor}
\usepackage[linesnumbered,ruled,vlined]{algorithm2e}
\usepackage{amsmath}
\usepackage{amssymb}
\usepackage{hyperref}
\hypersetup{
    colorlinks=true,
    linkcolor=blue,
    filecolor=magenta,      
    urlcolor=cyan,
    citecolor=blue,
}

\title{\LARGE \bf
Hierarchical Control Strategy for Moving A Robot Manipulator Between Small Containers
}

\author{Paolo Torrado$^{1}$, Boling Yang$^{2}$, and Joshua R. Smith$^{1, 2}$
\thanks{This work was supported by Amazon Inc through the UW+Amazon Science Hub.}
\thanks{$^{1}$ Electrical and Computer Engineering Department, University of Washington}%
\thanks{$^{2}$ Paul G. Allen School of Computer Science \& Engineering, University of Washington}
}

\begin{document}

\maketitle
\thispagestyle{empty}
\pagestyle{empty}

\begin{abstract}


In this paper, we study the implementation of a model predictive controller (MPC) for the task of object manipulation in a highly uncertain environment (e.g., picking objects from a semi-flexible array of densely packed bins). As a real-time perception-driven feedback controller, MPC is robust to the uncertainties in this environment. However, our experiment shows MPC cannot control a robot to complete a sequence of motions in a heavily occluded environment due to its myopic nature. It will benefit from adding a high-level policy that adaptively adjusts the optimization problem for MPC.

\end{abstract}

\section{Introduction}
Transferring objects between small containers is a popular robotic manipulation task, and it is particularly common in warehouse manipulation settings. It is a difficult control problem for an autonomous robot, requiring precise control and robust collision avoidance. The task complexity is further increased when the array of small containers is large because it heavily occludes the task space and limits the robot's maneuverability.


In this paper, we compared the performance of a MPC controller and a MPC-based hierarchical controller in a warehouse manipulation task. This task requires the robot to transfer objects between multiple containers and a shipping tote. MPC is an online control strategy that works well in stochastic environments. Yet, its sampling process and parameter-sensitive nature introduces uncertainty to its performance in complex environments. To overcome this weakness, we propose a MPC-based hierarchical controller that uses a policy that adaptively generates high-level strategies for the MPC algorithm to solve. Our experiment shows the hierarchical controller significantly outperformed the MPC controller in terms of success rate in task completion, control quality, and computational cost. We found that the hierarchical control strategy is effective in solving our particular manipulation task. In future research, we propose to improve the approach via reinforcement learning and curriculum learning.


\section{Sampling-Based Model Predictive Control}
MPC performs well in unstructured and dynamic environments requiring online adaptation~\cite{williams2017information}. It finds a locally optimal policy from the construction of multiple possible trajectories based on an approximate dynamics model \cite{bhardwaj2022storm}. Trajectories are discretized into particles representing future possible states of the robot and calculated a number of time steps into the future called a horizon. 
The optimal policy is obtained through the use of a loss function defined as follows:
\begin{align*}
\hat c(x_t,u_t)=\alpha_p \hat c_{pose} + \alpha_s \hat c_{stop} + \alpha_j \hat c_{joint} + \alpha_m \hat c_{manip} \\ + \alpha_c (\hat c_{self-coll} + \hat c_{env-coll})
\end{align*}

where $\hat c_{pose}$ penalizes distance to target pose, $\hat c_{stop}$ is the cost to stop for contingencies, $\hat c_{joint}$ is the joint limit avoidance, $\hat c_{manip}$ is the manipulability cost, $\hat c_{self-coll}$ self collision avoidance cost and $\hat c_{env-coll}$ is the environment collision cost. All the alpha terms are weight factors. 

\begin{figure}[t]
\centering
\includegraphics[width=0.38\textwidth]{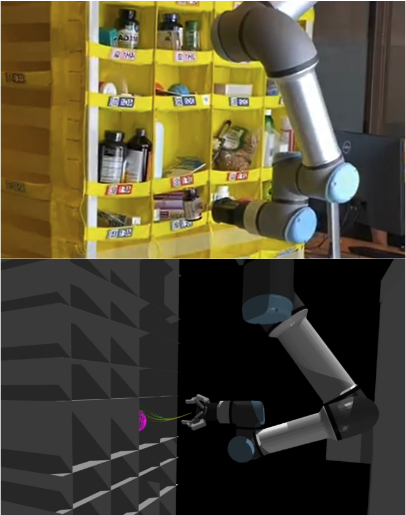}
\caption{Universal Robot UR16e depicted picking items from the array of bins in the real world and in simulation reaching a target. Qualitative results (videos) available at \href{run:https://clipchamp.com/watch/ONcvG3yN2Oc}{this link}.}
\label{fig:envs}
\end{figure}

\begin{figure*}[th!]
\centering
\includegraphics[width=0.96\textwidth]{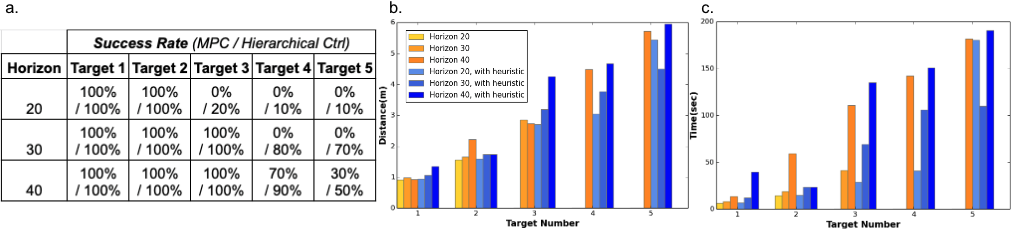}
\caption{Experimental Results based on 10 trials for each run. (a) Table presents the average success rate per 10 trials. (b) End effector traversed distance. (c) Time elapsed from start of simulation to target being reached.}
\label{fig:exp}
\end{figure*}

\section{Benchmarking Controller Performance}
We implemented a MPC controller on a Universal Robot UR16e and in simulation with IsaacGym \cite{liang2018gpu}. The design of the experiment is based on experience manipulating objects with the real robot. Benchmarking was  done on the simulation setup. The trajectory designed for the experiment represents successful and unsuccessful picks. 

The experiment consists of the robot reaching five waypoints. In waypoints one and two the robot reaches into a bin to simulate the grasping of an object and retrieving it to drop it in a shipping tote. Waypoints three, four and five involve reaching into a bin, failing to grasp an object and moving onto the next bin. 

The experiment was performed with three variations of the horizon parameter in the MPC controller. The number of particles for the trajectory calculation remained unchanged. Fig.\ref{fig:exp}.a shows all MPC controller settings had good performance reaching the first two waypoints. Success rate starts declining in the third waypoint. The controller does not overcome target three with a horizon of twenty and target 4 with a horizon of thirty. With a horizon of forty targets are reached with a low success rate. 

We proceeded with hierarchical control by adding a function to dictate global optimization. We choose to use a heuristic function as the high level policy in this experiment; planning and machine learning based approaches are also feasible policy alternatives.

We repeated the experiment with hierarchical control. Fig.\ref{fig:exp}.a. shows an improved target success rate. In the case of a horizon of twenty the controller is able to reach targets that were inaccessible. Comparing the runs with a horizon of forty shows the success rate is increased twenty percent for the last two targets. As presented in Fig.\ref{fig:exp}.b and c, the hierarchical controller with reduced horizon is more efficient in time and traversed distance than the MPC controller with longer horizon. Fig.\ref{fig:exp}.b and c show the hierarchical controller takes longer time and higher traversed distance in the trials with a horizon of forty.

MPC is a good low level controller for solving local optimization problems but has difficulties with finding global optimums. The last three targets of the experiment prove this point. The controller is unable to reason about navigating around the wall separating adjacent bins. A larger horizon for the controller increases the success rate of reaching waypoints between adjacent bins but at higher computational costs. The introduction of hierarchical control helps the robot reach a better global optimum while decreasing computational cost. In conclusion, the success rate of the robot is increased for all three settings of the horizon with the use of hierarchical control.

The heuristic function we use for hierarchical control adds additional waypoints to the trajectory of the robot. Naturally a greater number of waypoints increases the time and traversed distance required to reach targets. The reduction in performance is counterbalanced by greater number of original targets reached. The controller temporally sacrifices local optimum while improving the global optimum solution.


\section{Hierarchical Control}

While the heuristic policy effectively improved the robot's performance in our experiment, it will not scale well to a more diverse problem set the robot will encounter in realistic settings. The increase in complexity associated with densely packed container arrays will require more sophisticated manipulation skills. For example, other objects may occlude the target object requiring the robot to rearrange the container. Such non linear dynamics will increase the uncertainty in the performance of our hierarchical controller. 

To extend the functionality and generalizability of the hierarchical controller, we propose to use a reinforcement learning agent as the high-level policy and train the agent with curriculum learning strategies~\cite{DBLP:journals/corr/abs-2101-10382}. Curriculum learning is a training strategy that trains a machine learning model from easier data to harder data, which imitates the meaningful learning order in human curricula. We will be experimenting with two curriculum generation methods. The first method requires human experts to design oracles that generate training tasks. The second method is an autocurriculum approach~\cite{yang2022motivating, yang2022stackelberg}, where another RL agent is responsible for creating the curriculum. The robot manipulator and task generator are trained simultaneously to ensure that the generated tasks are reasonably challenging to the current robot policy. The robot and task generator are constantly improving their policies until reaching equilibrium.


\section{Conclusions}
We implemented an MPC controller in a UR16e robot and a simulation environment to study its feasibility as a low level controller for a hierarchical control strategy. The experiment consisted of having the robot traverse a series of waypoints representing the picking of a list of items. The MPC controller showed mixed performance in this task and was remarkably improved with the use of hierarchical control. The improved results gives us confidence a hierarchical control strategy combining the MPC controller and RL can be successful in reducing controller performance uncertainty.

\newpage
\bibliographystyle{IEEEtran}
\bibliography{ref.bib}






\end{document}